\documentclass[conference]{IEEEtran}
\usepackage{csquotes}

\ifCLASSINFOpdf
\else
\fi
\usepackage[pdftex]{graphicx}
\usepackage{enumitem}
\usepackage{amsfonts,amsthm,bm}
\usepackage{amsmath}
\usepackage[ruled,vlined]{algorithm2e}
\usepackage{subfigure}
\usepackage{float}
\usepackage{hyperref}
\begin{document}
%
\title{Stochastic Gradient Variance Reduction\\ by Solving a Filtering Problem}

\author{\IEEEauthorblockN{Xingyi Yang}
\IEEEauthorblockA{School of Electrical and Computer Engineering\\
University of California, San Diego\\
9500 Gilman Dr, La Jolla, CA 92093\\
Email: x3yang@eng.ucsd.edu}
}


%


\maketitle

\begin{abstract}
Deep neural networks (DNN) are typically optimized using stochastic gradient descent (SGD). However, the estimation of the gradient using stochastic samples tends to be noisy and unreliable, resulting in large gradient variance and bad convergence. In this paper, we propose \textbf{Filter Gradient Decent}~(FGD), an efficient stochastic optimization algorithm that makes the consistent estimation of the local gradient by solving an adaptive filtering problem with different design of filters. Our method reduces variance in stochastic gradient descent by incorporating the historical states to enhance the current estimation. It is able to correct noisy gradient direction as well as to accelerate the convergence of learning. We demonstrate the effectiveness of the proposed Filter Gradient Descent on numerical optimization and training neural networks, where it achieves superior and robust performance compared with traditional momentum-based methods. To the best of our knowledge, we are the first to provide a practical solution that integrates filtering into gradient estimation by making the analogy between gradient estimation and filtering problem in signal processing. \footnote{The code is provided in \url{https://github.com/Adamdad/Filter-Gradient-Decent}}  
\end{abstract}


%
\IEEEpeerreviewmaketitle

\section{Introduction}
\label{sec:intro}

Among common techniques to solve the
optimization problem, stochastic gradient descent (SGD) is one of the most common practice nowadays. It aims to
minimize the an objective function $J(\bm{\theta}_t;\cdot)$ by taking the gradient computed from the randomly sampled data iteratively. Consider training a parametric model $f(\cdot;\bm{\theta})$ parameterized by $\bm{\theta}\in \mathbb{R}^d$ over a finite set of training data $\mathcal{D}=\{(\mathbf{x}_i,y_i)\}_{i=1}^N$. At iteration $t$, a subset of the dataset $\mathcal{D}_t \subseteq \mathcal{D}$ is randomly sampled. Denote $\bm g_t=\nabla_{\bm{\theta}}J(\bm{\theta}_t;\mathcal{D}_t) $ as the observed gradient, we minimize the objective by updating the weight $\bm{\theta}$ in the down-hill direction 
\begin{equation}
    \bm{\theta}_{t+1} = \bm{\theta}_{t}-\eta \bm g_{t}\label{eq:gd}
\end{equation}
The learning rate $\eta$ determines the stepsize heading to the minimum.

A critical problem of SGD is that it is highly ``greedy''. On the one hand, the sampled data points $\mathcal{D}_t$ might not well represent the overall data statistics, e.g. class-biased and noise-corrupted, which leads to undesirable error for the gradient estimation. Meanwhile, the stochastic gradient only considers local information for the error surface, therefore might spend much time bouncing around ravines with poorly conditioned Hessian matrix. Consequently, the estimated gradient has a large variance~\cite{wang2013variance}, leading to slower convergence and worse performance.
This phenomenon is most severe for the non-smooth and non-convex optimization problems, like training deep neural networks. 
Existing problems in stochastic gradient descent forces us to think about how to alleviate the randomness in gradient calculation and solve the problem of poor convergence. 

In the theory of stochastic processes, the filtering problem is a model for state estimation problems to establish a ``best estimate'' for the true value of some system from an incomplete, potentially noisy set of observations on that system. This is closely related to our problem: \textit{Given observations $\{\bm g_i\}$ for $0 \leq i \leq t$, what is the best estimate $\hat{\bm g}_{t|t}$ of the true state $\hat{\bm g}_t$ of the system based on those observations?}

\begin{figure}
    \centering
    \includegraphics[width=\linewidth]{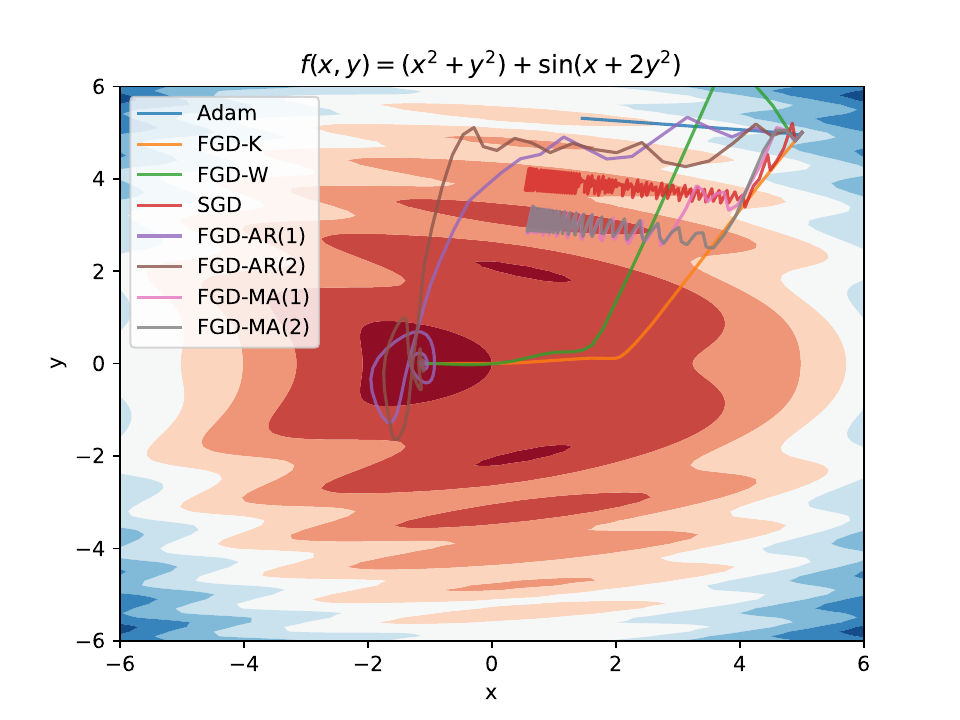}
    \caption{Illustration of Filter Gradient Decent~(FGD) for a non-convex optimization problem. Using common filtering techniques to ``denoise'' the gradient observation, FGD is able to reduce the variance of the gradient as well as to get rid of the local minimum in SGD.}
    \label{fig:illustration}
\end{figure}

Inspired by the successful application for filters in signal denoising and reconstruction, we introduce a general framework called \textbf{Filter Gradient Decent}~(FGD), a remedy to the noisy gradient with the filters to reduce variance and make consistent estimation of gradient in stochastic optimization. It takes the stochastic gradient as a special signal and solving the filtering problem or denoising problem to estimate the true gradient direction under the noisy observations. As a simple illustration in Figure~\ref{fig:illustration}, FGD is able to reduce the variance of the gradient as well as to get rid of the local minimum in SGD. This general framework can be applied to any filtering techniques.

Our contributions are summarized as follows:
\begin{itemize}[noitemsep]
    \item We propose Filter Gradient Decent~(FGD), which incorporates signal filtering into stochastic gradient descent to reduce the noise and randomness of gradient estimation.
    \item We show that our framework can be applied to any filters. In this paper, we compare four kinds of filters including (1) Moving-average~(MA) filter (2) Autoregressive~(AR) filter (3) Kalman filter (4) Wavelet filter.
    \item Experiments verify that Filter Gradient Decent is effective in gradient variance reduction and finding the global minimum for non-convex optimization problem.
    \item We train deep neural networks with Filter Gradient Decent under various datasets, which demonstrates superior performance and fast convergence.
\end{itemize}
\section{Background}
\subsection{Stochastic Gradient Descent} Stochastic gradient descent~(SGD) is a way to minimize an objective function by updating the parameters in the opposite direction of the gradient. Vanilla SGD, however, does not guarantee good convergence, either suffering from the large variance of the gradient or trapped in suboptimal local minima. Momentum~\cite{qian1999momentum} helps accelerate SGD by adding a ratio of $\gamma$ of the update vector of the past time step to the current update vector. It also reduces the gradient variance by keeping a running average of past observations. Adagrad~\cite{duchi2011adaptive} uses a different learning rate for every parameter $\theta_i$ at every time step $t$. Adadelta~\cite{zeiler2012adadelta} and RMSprop~\cite{hinton2012neural} are two extensions of Adagrad that seeks to restrict its aggressive learning rate. Adaptive Moment Estimation (Adam)\cite{kingma2014adam} incorporates adaptive learning rates with a decaying average of past gradients similar to momentum. Later variants like AdaMax\cite{kingma2014adam} and Nadam`\cite{dozat2016incorporating} incrementally extends the framework of Adam.\\
\subsection{Variance Reduction in Gradient Descent} SGD uses the noisy gradient estimated from random
data samples. This method introduces a large variance to gradient estimation, which is harmful to the model in terms of convergence. Apparently, averaging is a simple yet off-the-shelf solution. Averaged Stochastic Gradient Descent~(ASGD)~\cite{bottou2012stochastic} is the simplest method of variance reduction by keeping an average of some set of the weight vectors. Stochastic Averaged Gradient~(SAG)~\cite{schmidt2017minimizing} builds on the ASGD method by keeping a moving average of $n$ past gradients for updates. \cite{wang2013variance} builds a variance reduction technique, which makes use of control variates to augment the noisy gradient and thereby reduce its variance. \cite{miller2017reducing} proposes to control the variance of the reparameterization gradient estimator in Monte Carlo variational inference (MCVI). 

Averaging is equivalent to a simple lowpass filter in signal processing. It removes the high frequency component as well as the noise in signals. Recently, Kalman Optimizer~\cite{9414588} show that kalman filter can be applied to gradient decent to reduce the variance of the gradient. Our FGD is partially inspired by the averaging techniques and Kalman Optimizer: If an average lowpass filter works for gradient variance reduction, sophisticated filters may work better.

\subsection{Filters}
In signal processing, a filter is a device or process that removes some unwanted components or features from a signal. We introduce four types of filters used in this paper and their basic concepts.\\
\textbf{Moving-average Filter}. 
In time series analysis, the moving-average~(MA) model specifies that the output variable depends linearly on the current and various past values of a stochastic term. $MA(q)$ refers to the moving average model of order q, where the output $\mathbf{y}_t$ is a weighted linear combination of $\mathbf{x}_{t}$ and $q$ past input $\{\mathbf{x}_{t-1},\dots,\mathbf{x}_{t-p}\}$ 
\begin{equation}
    \mathbf{y}_t=\mathbf{x}_{t} + \sum_{i=1}^q a_i \mathbf{x}_{t-i}
\end{equation}
\\
\textbf{Autoregressive Filter}. An autoregressive~(AR) model is a type of random process that the output variable depends linearly on its own previous values and on a stochastic term. $AR(p)$ indicates an autoregressive model of order p, where the output $\mathbf{y}_t$ is a weighted linear combination of input $\mathbf{x}_{t}$ and $p$ past output value $\{\mathbf{y}_{t-1},\dots,\mathbf{y}_{t-p}\}$
\begin{equation}
    \mathbf{y}_t=\mathbf{x}_{t} + \sum_{i=1}^p b_i \mathbf{y}_{t-i}
\end{equation}\\
\textbf{Kalman Filter}. 
The Kalman filter~\cite{kalman1960new} addresses the general problem of trying to estimate the state $\mathbf{x} \in \mathbb{R}^n$ of a discrete-time controlled process that is governed by the linear stochastic difference equation with a measurement $\mathbf{y} \in \mathbb{R}^m$
\begin{equation}
    \mathbf{x}_t =  \mathbf{F}_{t-1} \mathbf{x}_{t-1}+\mathbf{I}_{t-1} \mathbf{u}_{t}+\mathbf{w}_{t-1}, \quad \mathbf{y}_{t}=\mathbf{C}_{t-1} \mathbf{x}_{t}+\mathbf{v}_{t}
\end{equation}
$\mathbf{F}_{t}\in \mathbb{R}^{n\times n}$ is the state transition matrix which relates the state to the previous step, $\mathbf{I}_{t} \in \mathbb{R}^{n\times k}$ is the control matrix that relates the optional control input $\mathbf{u} \in \mathbb{R}^k$ to current state, and $\mathbf{C}_{t}\in \mathbb{R}^{m\times n}$ is measurement matrix. The random variables $\mathbf{w}_{t} \sim \mathcal{N}(0,Q)$ and $\mathbf{v}_{t} \sim\mathcal{N}(0,R)$ represent the process and measurement noise respectively. They are assumed to be independent of each other, white, and with normal probability distributions.

The Kalman filter solves the problem in two steps: the prediction step, where the next state of the system is predicted given the previous measurements, and the update step, where the current state of the system is estimated given the measurement at that time step.
The steps translate to equations as follows
\begin{itemize}
    \item Prediction Step
    \begin{align}
        \hat{\mathbf{x}}_{t|t-1} = \mathbf{F}_{t-1}\mathbf{x}_{t-1} + \mathbf{I}_{t}\mathbf{u}_{t}\\
         \hat{\mathbf{P}}_{t|t-1} = \mathbf{F}_{t-1}\mathbf{P}_{t-1|t-1}\mathbf{F}_{t-1}^T + Q
    \end{align}
    \item Update Step
    \begin{align}
        \hat{\mathbf{y}}_t= \mathbf{y}_t-\mathbf{C}  \hat{\mathbf{x}}_{t|t-1}\\
        \mathbf{K}_{t} = \hat{\mathbf{P}}_{t|t-1}\mathbf{C}^T(\mathbf{C} \hat{\mathbf{P}}_{t|t-1} \mathbf{C}^T + R)^{-1}\\
        \hat{\mathbf{x}}_{t|t}= \hat{\mathbf{x}}_{t|t-1} +\mathbf{K}_{t}\hat{\mathbf{y}}_t\\
        \mathbf{P}_{t|t} = (\mathbf{I}-\mathbf{K}_t\mathbf{C})\hat{\mathbf{P}}_{t|t-1}
    \end{align}
\end{itemize}
\textbf{Wavelet Filter}. A wavelet is a mathematical function used to divide a given function or continuous-time signal into different scale components. A wavelet transform is the representation of a function by wavelets. By shifting and scaling a mother wavelet function $\Psi(t)$, any function $f(t)$ can decomposed into cofficents with multi-resolution wavelets 
\begin{equation}
    \Psi_{a,b}(t) =\frac{1}{\sqrt{b}} \Psi(\frac{t-a}{b}), \gamma_{a,b} = \int f(t) \Psi_{a,b}(t) dt
\end{equation}
where $a$ and $b$ are translation and scaling coefficients, $\gamma_{a,b}$ is the wavelet coefficients correspond to wavelet $\Psi_{a,b}(t)$.
A discrete wavelet transform~(DWT) is any wavelet transform for which the wavelets are discretely sampled. There are three basic steps to filtering signals using wavelets. (1) Decompose the signal using the DWT.
\begin{equation}
    \Psi_{j,k}(t) =\frac{1}{\sqrt{2^{j}}} \Psi(\frac{t-k2^j}{2^j}), \gamma_{j,k} = \sum_{t} f(t) \Psi_{j,k}(t) 
\end{equation}
(2) Filter the signal in the wavelet space using thresholding operation $T$.
\begin{equation}
    \gamma'_{j,k}=T(\gamma_{j,k})
\end{equation}
(3) Invert the filtered wavelet coefficients to reconstruct the filtered signal using the inverse DWT.
\begin{equation}
    f'(t) = \sum_i\sum_k \gamma'_{j,k} \Psi_{j,k}(t) 
\end{equation}
The filtering of signals using wavelets is based on the idea that as the DWT decomposes the signal into details and approximation parts. At some scale the details contain mostly the insignificant noise and can be removed or zeroed out using thresholding without affecting the signal.
\begin{figure*}
    \centering
    \includegraphics[width=\linewidth]{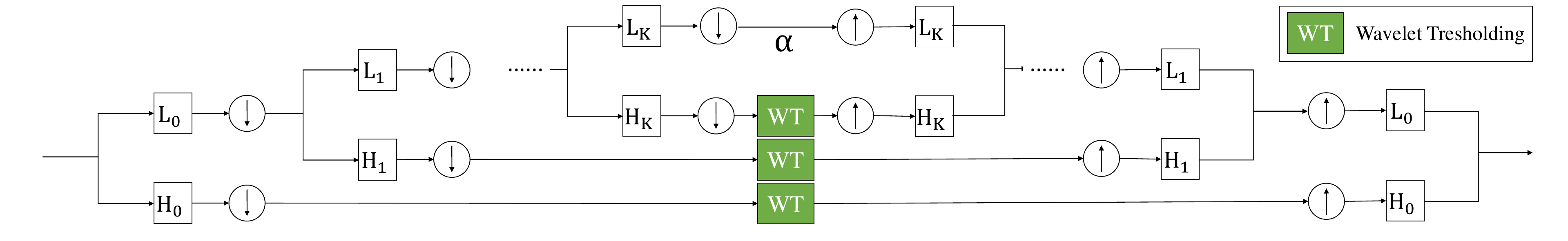}
    \caption{Pipeline for Wavelet Gradient Filtering for Gradient Decent. (1) Forward DWT to get multi-resolution representation of the gradients (2) Thresholding and amplification on wavelet domain (3) Inverse DWT to spatial domain.}
    \label{fig:WGD}
\end{figure*}
\section{Gradient Estimation with a Filtering Problem}
In this paper, we make a bold analogy between gradient and high-dimensional signal and reliably estimate gradient by solving a filtering problem. Given the current and past noisy gradient observation $\{\bm g_i\}_{i=0}^t$, we estimate the true state of the current gradient $\hat{\bm g}_{t|t}$ using filter $F$
\begin{equation}
    \hat{\bm g}_{t|t}= F(\bm g_t,\bm g_{t-1},\dots, \bm g_{0})
\end{equation}
The estimated gradient then is utilized to update the model weight $\bm \theta$ via Equation~\ref{eq:gd}. Algorithm~\ref{alg:FGD} shows the overall framework of Filter Gradient Decent.
\begin{algorithm}
\SetAlgoLined
\label{alg:FGD}
\KwResult{Network Parameter $\bm \theta$}
 Initialize the Network Parameter with $\bm \theta_0$ and Filter $F$ with parameter $\Gamma$;\\
 \For{$t=0,\dots,T-1$}{
  Compute Gradient:\\
  \quad$\bm g_t=\nabla_{\bm{\theta}}J(\bm{\theta}_t;\mathcal{D}_t) $\;
  Filtering Step:\\
  \quad$\hat{\bm g}_{t|t}= F(\bm g_t,\bm g_{t-1},\dots, \bm g_{0})$\;
  Gradient Descent:\\
  \quad $\bm{\theta}_{t+1} = \bm{\theta}_{t}-\eta \hat{\bm g}_{t|t}$\;
 }
\caption{Filter Gradient Decent}
\end{algorithm}
The different filter selection models the various parametric dependencies between the true state and the gradient computed from randomly sampled data. It works as a low-pass filter that removes the high-frequency components~(randomness and noise) in the gradient observations and enhances the low-frequency components~(consistent direction of the gradient). Gradient estimation with filter gets rid of the drawbacks of local gradient and data randomness by considering the historical states of the gradient, thus become more consistent and robust. We select four well-known filters (1) MA filter (2) AR filter (3) Kalman filter and (4) Wavelet filter as $F$ and would like to study how different filters affect the gradient estimation. 
\subsection{Solution 1: Moving-average for Gradient Decent}
When assuming that the observed and true gradients follow a moving-average model, we can approximate the true gradient $\hat{\bm g}_{t|t}$ using an $MA(q)$ model
\begin{align*}
    \hat{\bm g}_{t|t} = \bm g_t + \sum_{i=1}^q a_i \bm g_{t-i}
\end{align*}
$\{ \bm g_{t-i}\}_{i=1}^q$ is $q$ past stochastic gradients and $a_i$ is the weight for historical gradient $\bm g_{t-i}$. We call this method Filter Gradient Decent with Moving-average filter~(FGD-MA). In the following paper, we call the FGD-MA model with order $q$ as FGD-MA(q). It keep a weighted averaging of $q$ past gradient observations. 
\subsection{Solution 2: Autoregresive for Gradient Decent}
When the autoregressive model best describes the relation between the observed and true gradients, we may also estimate the true gradient $\hat{\bm g}_{t|t}$ using an $AR(p)$ model
\begin{align*}
    \hat{\bm g}_{t|t} = \bm g_t + \sum_{i=1}^p b_i \hat{\bm g}_{t-i|t-i}
\end{align*}
The $b_i$ is the weight corresponds to historical gradient estimation $\hat{\bm g}_{t-i|t-i}$. We call this model Filter Gradient Decent with Autoregresive filter~(FGD-AR). Interestingly, FGD-AR with order 1 turns out to be the momentum method that is commonly used in stochastic gradient decent
\begin{align*}
    \bm v_t = \bm g_t + \gamma \bm v_{t-1}
\end{align*}
This shows that the momentum method is one special case for FGD-AR. FGD-AR extends the current momentum term to allow longer time dependency for gradient estimation.  In the following paper, we call the FGD-AR model with order $p$ as FGD-AR(p). We also call momentum method as FGD-AR(1).
\subsection{Solution 3: Kalman filter for Gradient Decent}
 To reduce the variance of the gradient, we adopt a Kalman filter~\cite{9414588} to estimate the true state of gradient upon a filtering problem. We assume the gradient follows a linear scholastic model
\begin{equation}
\begin{split}
    \hat{\bm g}_t=\mathbf{F}_{t-1} \hat{\bm g}_{t-1}+\mathbf{w}_{t-1}, \quad
    \bm g_{t}=\mathbf{C}_{t-1} \hat{\bm g}_{t}+\mathbf{v}_{t-1}
\end{split}
\end{equation}
where $\bm g_{t}$ is the observed gradient and $\hat{g}_{t}$ refers to real gradient. We adopt the Kalman filter to give the minimum mean-square estimate of real gradient $\hat{\bm g}_{t|t}$ given previous state $\hat{\bm g}_{t-1|t-1}$ and observation $\bm g_{t}$ at time $t$.
\begin{equation}
    \hat{\bm g}_{t|t}= KF(\bm g_{t},\hat{\bm g}_{t-1|t-1})
\end{equation}

By assuming that the gradient for each parameter  is independent, we need to maintain a Kalman filter for each parameter. To reduce the computational burden, we simplified our method to a scalar Kalmen filter: $\mathbf{F}_{t} = \gamma I$, $\mathbf{C}_{t} = cI$, $Q = \sigma_Q^2 I$, $R = \sigma_R^2 I$, $\mathbf{P}_0 = p_0 I$, where $I$ is the identity matrix. This design cancels the time-consuming matrix multiplication, with only a scalar Kalman filter per parameter. We call this model Filter Gradient Descent with Kalman filter~(FGD-K).

\subsection{Solution 4: Wavelet Gradient Decent}

Wavelet Filtering is widely adopted in signal processing to remove certain frequency components of the signal. In order to better eliminate gradient noise, we consider the multi-resolution wavelet filter that soft-thresholds highpass coefficients at each level. We also intend to boost the low-frequency component of the gradient by amplifying the lowpass coefficient by a hyper-parameter of $\alpha$. Figure~\ref{fig:WGD} explains the overall pipeline for our proposed Filter Gradient Descent with Wavelet filter~(FGD-W). The procedure can be summarized into three steps
\begin{itemize}
    \item \textit{Forward DWT} We first apply K-level  wavelet transformation for $L$ historical gradients to get multi-resolution representation of the gradient
    \begin{equation}
        (\bm L_K, \bm H_K, \dots, \bm H_0)= DWT(\bm g_{t}, \dots,\bm g_{t-L+1})
    \end{equation}
    $\bm L_i$ and $\bm H_i$ are the lowpass and highpass coefficient vectors at level $i$.
    Particularly, we assume independence for different elements of the gradient and maintain a 1D DWT for each parameter. In terms of implementation, we keep a $L$-length queue for each parameter to store $L$ historical gradient observations. Every time a new gradient $\bm g_{t}$ is computed, it is pushed in the queue while $\bm g_{t-L}$ is popped out.    
    \item \textit{Filter in wavelet domain} Soft-thresholding is performed on all highpass components $\bm H_i$ at each level with operation $T$ and amplify the lowpass band by a coefficient of $\alpha\geq 1$
    \begin{equation}
    \bm H'_i= T(\bm H_i),\quad
    \bm L'_K= \alpha \bm L_K 
\end{equation}
    where $\bm H'_i$ and $\bm L'_K$ is the filtered highpass and low pass coefficients in wavelets domain. For simplicity, we adopt the same threshold on all levels. 
    \item \textit{Inverse DWT} The filtered coefficients in wavelet domain are converted back to spatial domain using inverse DWT
    \begin{align*}
        \bm g_{t|t}, \dots,\bm g_{t-L+1|t} = IDWT(\bm L'_K, \bm H'_K, \dots, \bm H'_0)
    \end{align*}
\end{itemize}

\begin{figure*}[t]
    \centering
    \subfigure[Function Value surface with contour]{\includegraphics[width=0.32\linewidth]{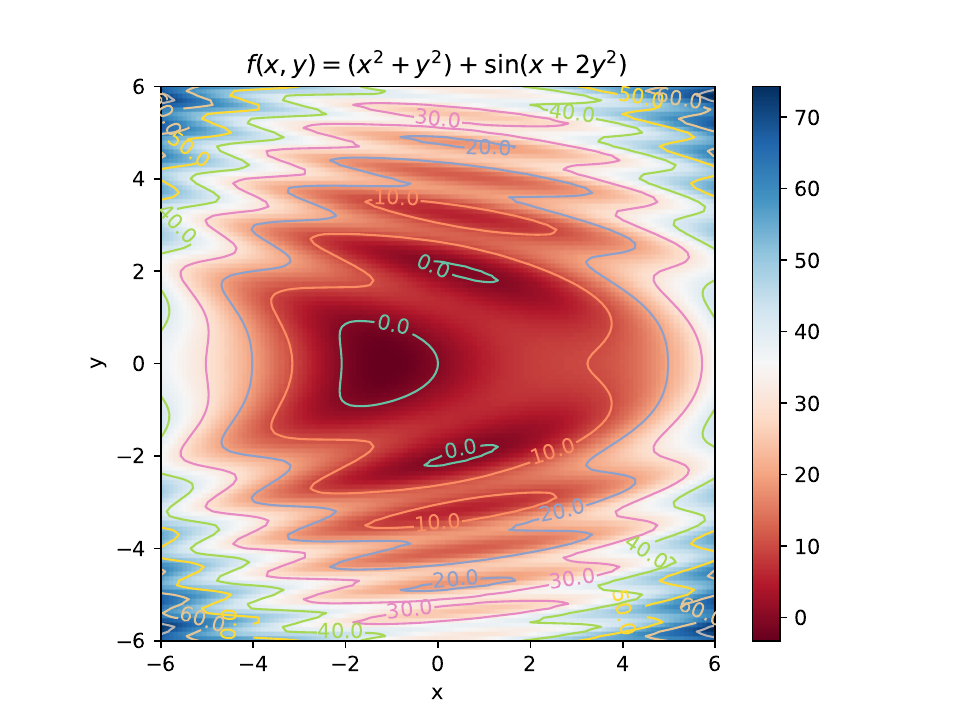}\label{fig:surface}}
    \subfigure[Learning rate 0.01]{\includegraphics[width=0.32\linewidth]{fig/NonConvex_0.01_output.pdf}\label{fig:traj0.01}}
    \subfigure[Learning rate 0.03]{\includegraphics[width=0.32\linewidth]{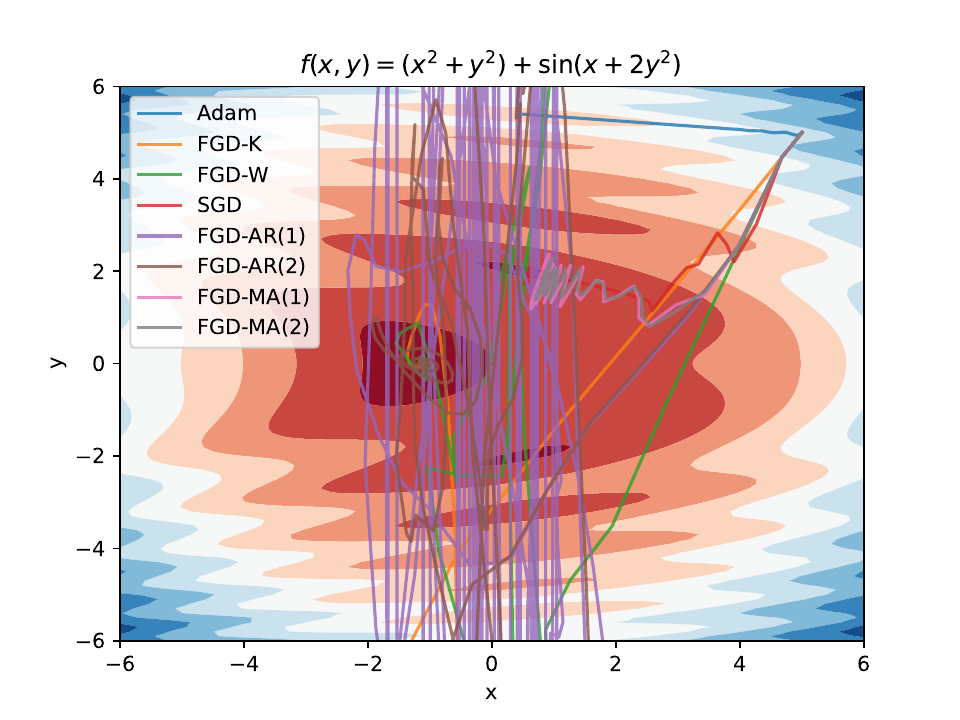}\label{fig:traj0.03}}
     \subfigure[$f(x,y)$, learning rate 0.01]{\includegraphics[width=0.45\linewidth]{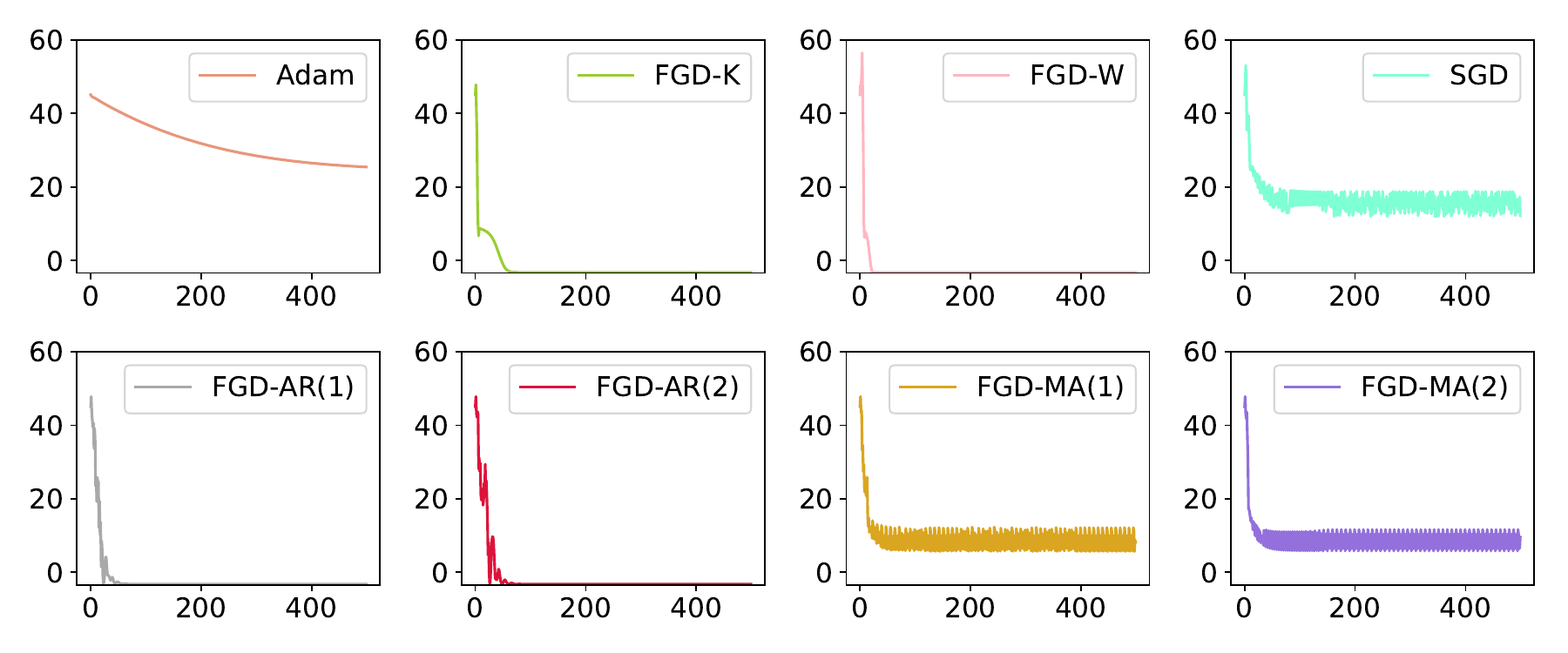}\label{0.01value}}
    \subfigure[$f(x,y)$, learning rate 0.03]{\includegraphics[width=0.45\linewidth]{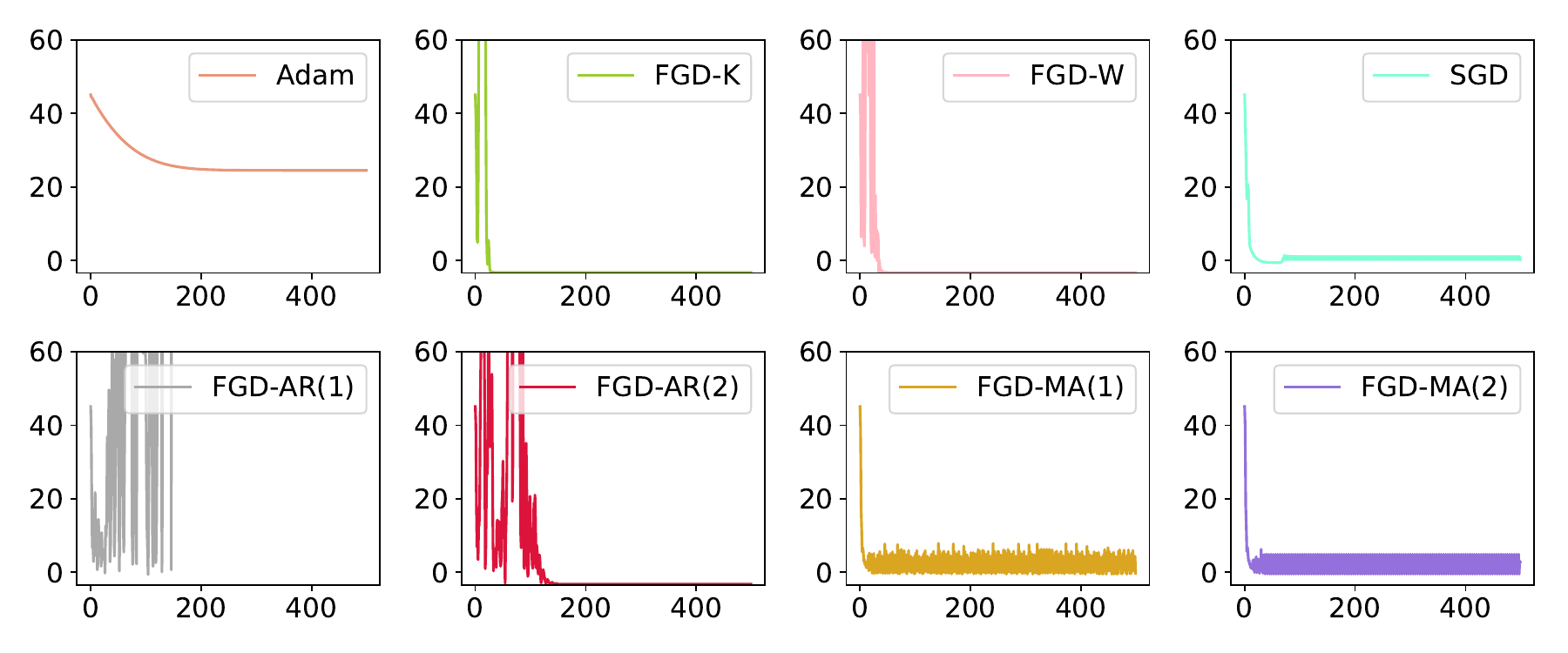}\label{0.03value}}
    \subfigure[$|\nabla_xf(x,y)|^2$, learning rate 0.01]{\includegraphics[width=0.45\linewidth]{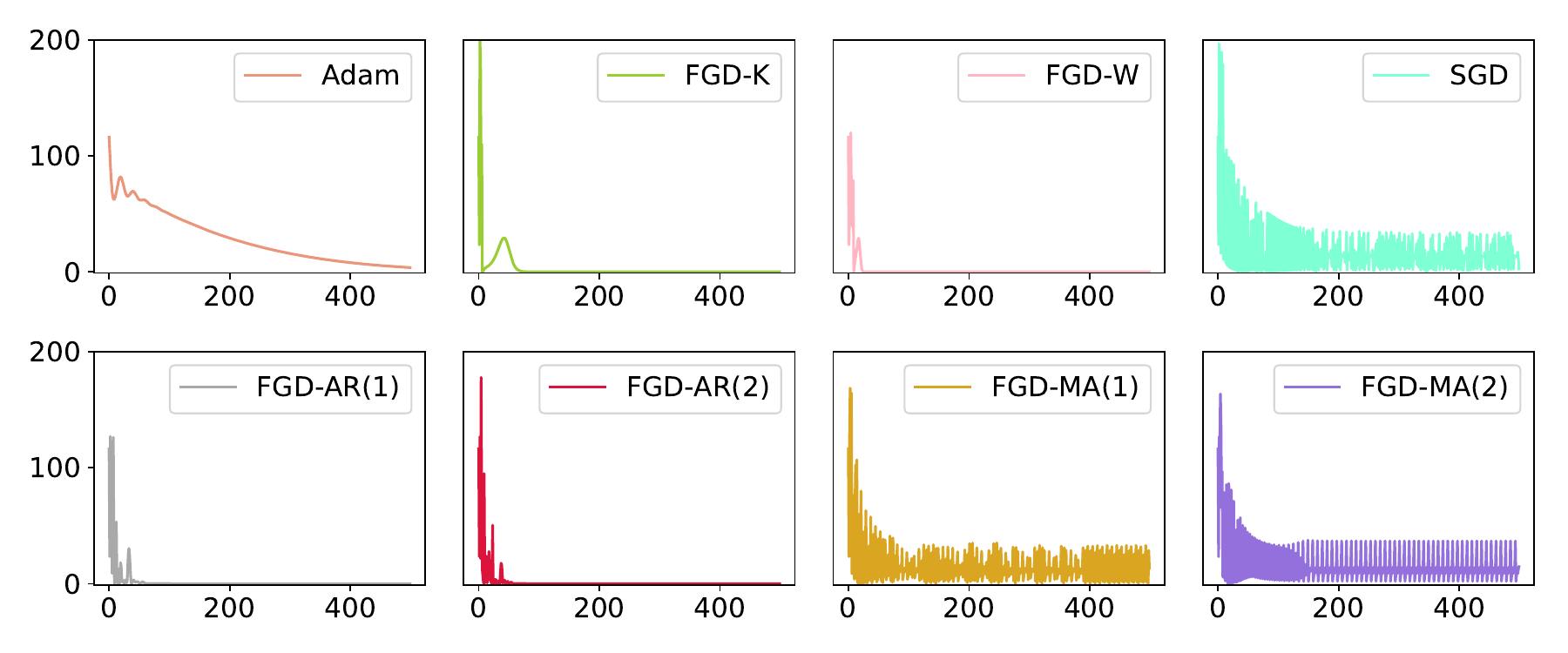}\label{0.01x var}}
    \subfigure[$|\nabla_xf(x,y)|^2$, learning rate 0.03]{\includegraphics[width=0.45\linewidth]{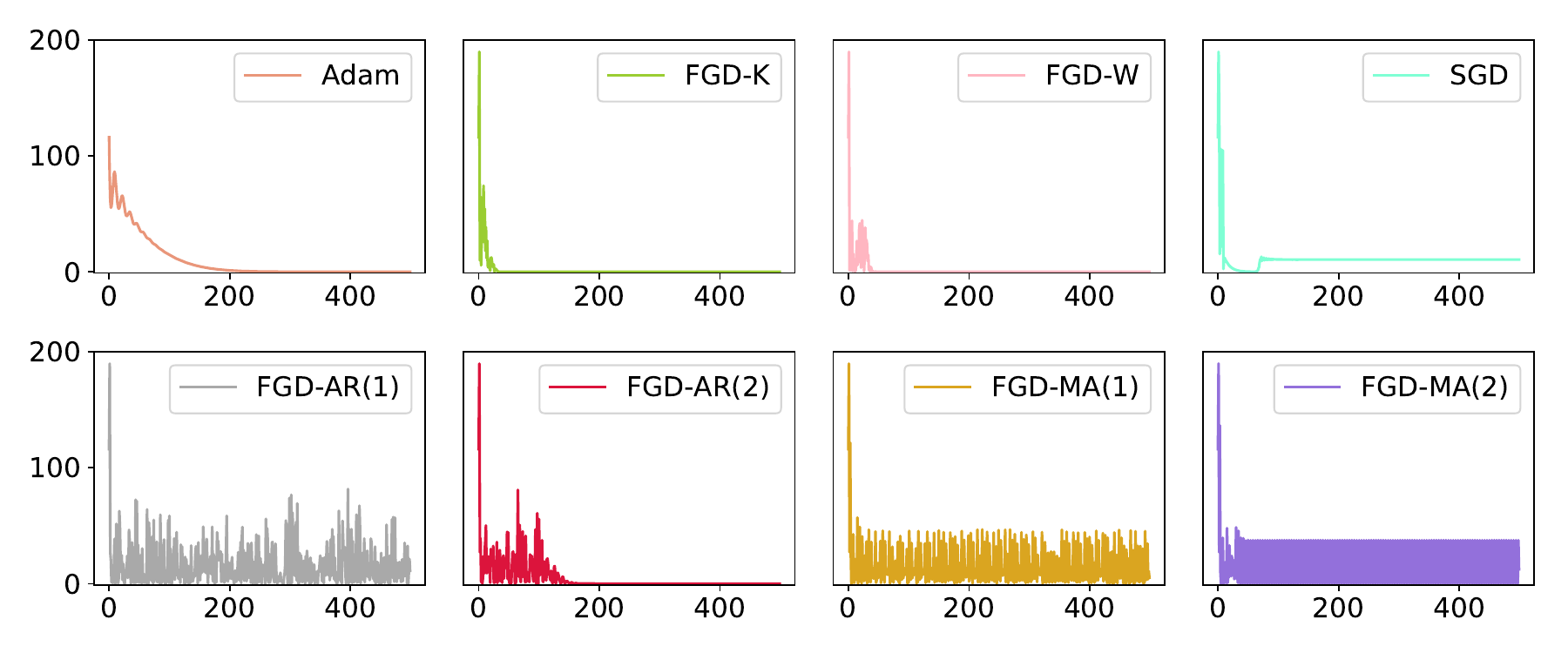}\label{0.03x var}}
    \subfigure[$|\nabla_yf(x,y)|^2$, learning rate 0.01]{\includegraphics[width=0.45\linewidth]{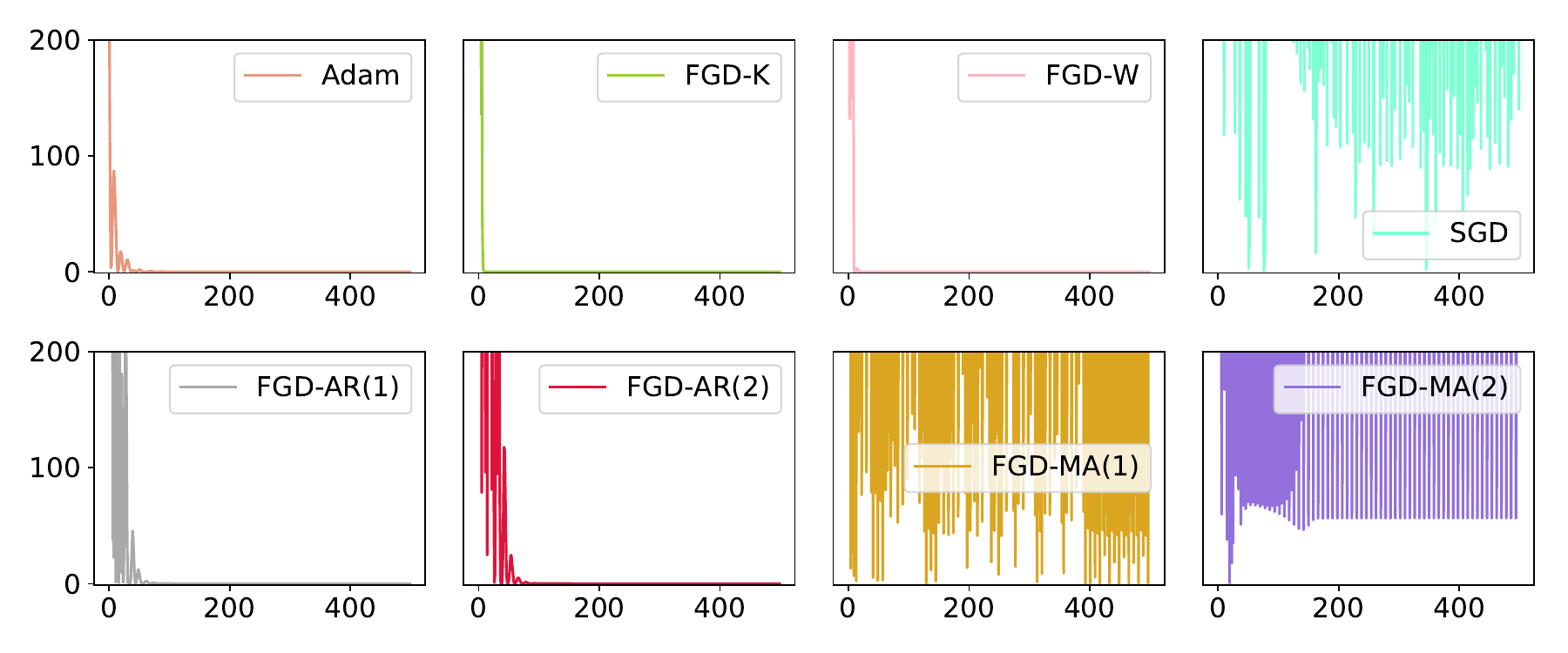}\label{0.01y var}}
    \subfigure[$|\nabla_yf(x,y)|^2$, learning rate 0.03]{\includegraphics[width=0.45\linewidth]{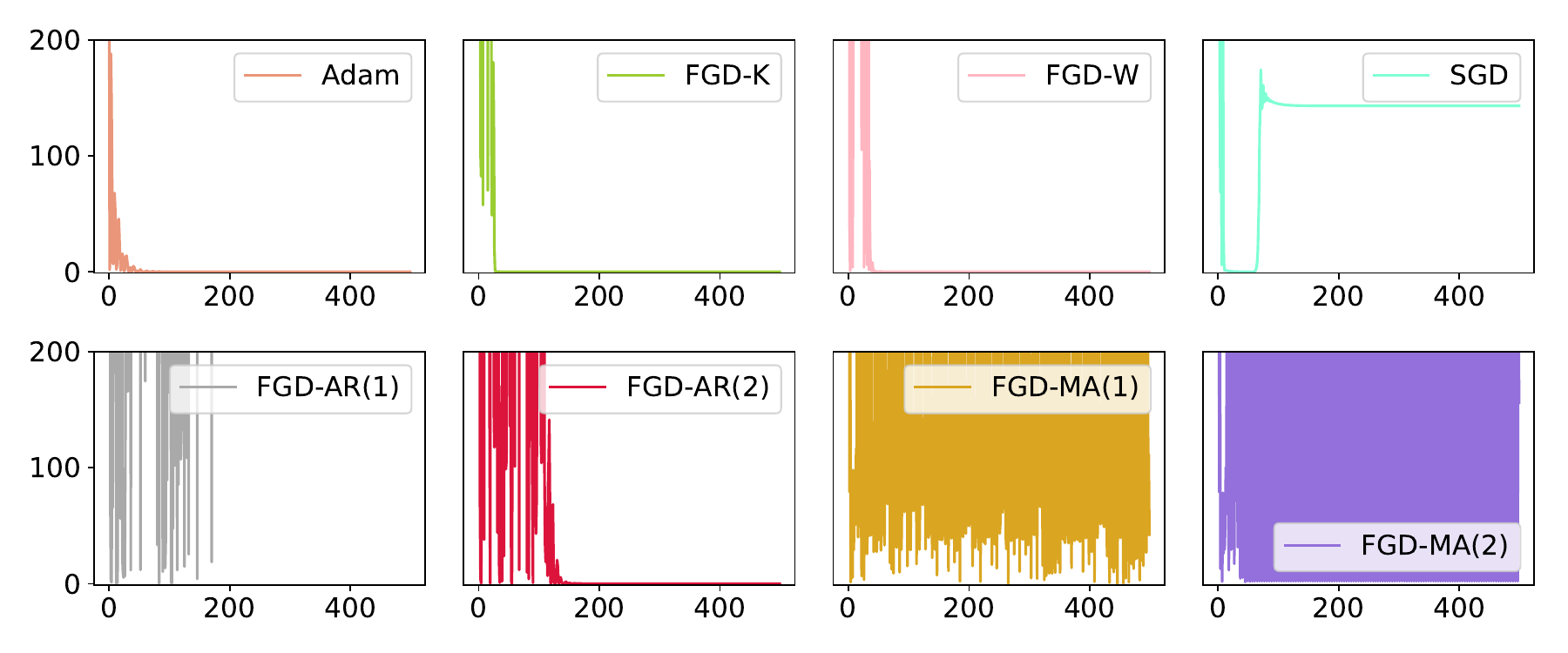}\label{0.03y var}}
    \caption{Plot of Value surface of $f(x,y)$ with contour, the function value, gradient variance for non-convex function minimization}
    \label{fig:nonconvex}
\end{figure*}
\begin{figure*}[h]
    \centering
    \subfigure[Batch-size 64, learning rate 0.001]{\includegraphics[width =\linewidth]{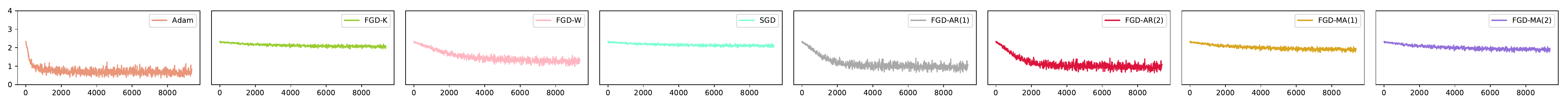}}
    \subfigure[Batch-size 64, learning rate 0.01]{\includegraphics[width =\linewidth]{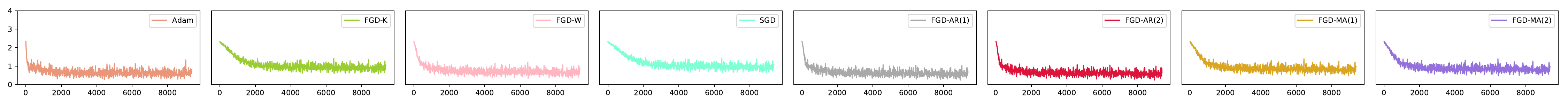}}
    \subfigure[Batch-size 64, learning rate 0.1]{\includegraphics[width =\linewidth]{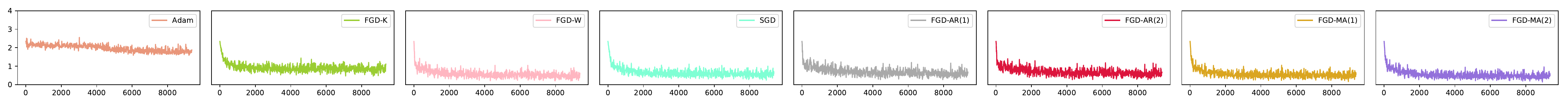}}
        \subfigure[Batch-size 16, learning rate 0.001]{\includegraphics[width =\linewidth]{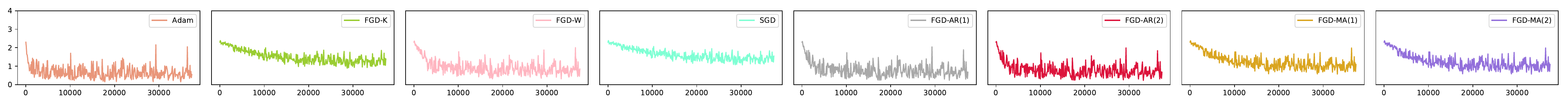}}
    \subfigure[Batch-size 16, learning rate 0.01]{\includegraphics[width =\linewidth]{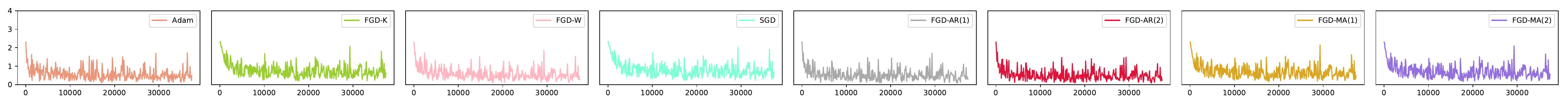}}
    \subfigure[Batch-size 16, learning rate 0.1]{\includegraphics[width =\linewidth]{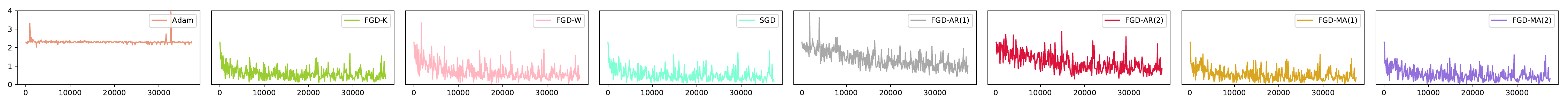}}
    \subfigure[Batch-size 4, learning rate 0.001]{\includegraphics[width =\linewidth]{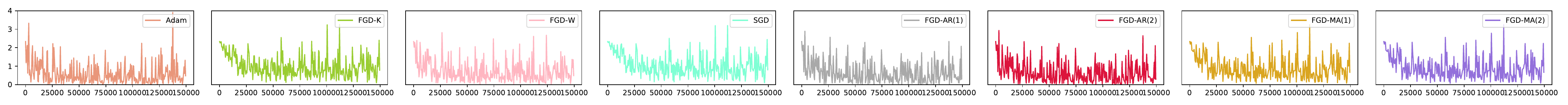}}
    \subfigure[Batch-size 4, learning rate 0.01]{\includegraphics[width =\linewidth]{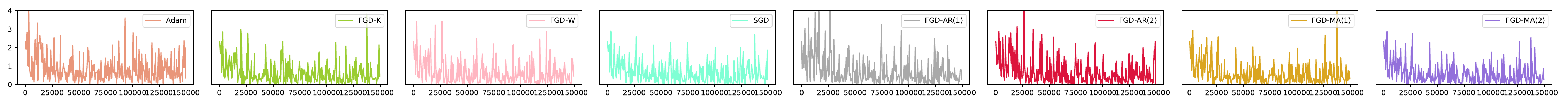}}
    \subfigure[Batch-size 4, learning rate 0.1]{\includegraphics[width =\linewidth]{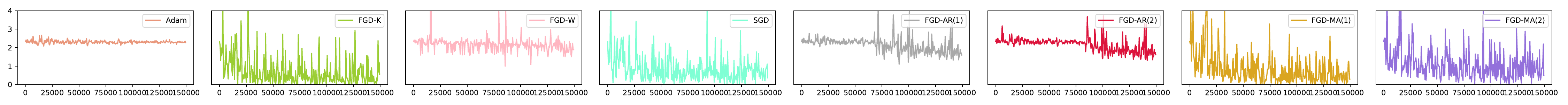}}
    \caption{Training loss on MNIST with MLP(x-axis: iteration; y-axis: loss)}
    \label{fig:MNIST training loss}
\end{figure*}
\section{Experiment}
To empirically evaluate the proposed method, we first show a numerical problem for non-convex optimization. We also investigated its performance in optimizing different popular machine learning models, including multi-layer fully connected neural networks and deep convolutional neural networks. Using various models and datasets, we demonstrate FGD with various filter designs can efficiently solve the numerical optimization and image classification problems. 

\begin{table*}[t]
    \centering
        \caption{Test Accuracy on MNIST with MLP with different hyper-parameter setting}
    \begin{tabular}{l|c|c|c|c|c|c|c|c}
    \hline
         Setting & Adam & FGD-K &FGD-W & SGD  & FGD-AR(1) &FGD-AR(2) & FGD-MA(1)  & FGD-MA(2)  \\
         \hline
         Batch 4\quad lr 0.001 &91.55 &85.41 & 90.16& 86.79& 90.60&90.93 &88.96 & 88.92\\
          Batch 16\quad lr 0.001 & 89.95& 80.59& 87.58 & 77.90&88.85 & 88.87& 84.38&84.38\\
           Batch 64\quad lr 0.001 & 88.55&36.79 &81.02 & 35.86&85.89 &85.88 & 50.63& 50.61\\\hline
            Batch 4\quad lr 0.01 &85.55 &91.41 & 91.48& 90.41& 91.18& 90.42&91.34 &90.73 \\
             Batch 16\quad lr 0.01 &90.71 & 89.09& 91.46&88.71 &91.66 &91.61 &89.68 &89.51 \\
              Batch 64\quad lr 0.01 & 87.86& 86.22& 89.26&85.96 &90.22 &90.16 &89.41 & 89.41\\\hline
              Batch 4\quad lr 0.1 & 11.52&91.29 & 20.53&88.07 &26.52 &26.16 &91.48 & 85.94 \\
             Batch 16\quad lr 0.1 & 12.03& 90.64& 89.54&91.77 & 71.81&79.36 & 92.13&92.07 \\
              Batch 64\quad lr 0.1 & 30.31& 81.80& 90.94& 90.18&89.67 &89.14 & 91.15& 91.23\\\hline
              Mean & 65.34 & 81.47	& 81.33&	81.74&	80.71& 81.39& \textbf{85.46}& 84.76\\\hline

         \hline
    \end{tabular}

    \label{tab:MNIST}

\end{table*}

\subsection{Study 1: Non-convex optimization}

We experiment with the scenario of optimizing a well designed non-convex problem using (1) SGD (2) Adam~\cite{kingma2014adam} (3) FGD-MA with order 1 and 2 (4) FGD-AR with order 1~(momentum) and 2 (5) FGD-K and (6) FGD-W. In this experiment, we intend to minimize a non-convex function
\begin{equation}
    f(x,y) = (x^2+y^2)+5\sin(x+y^2)
\end{equation}
\textbf{Experiment Setting}. For each experiment, we select a learning rate from $\eta=\{0.01,0.03\}$ to minimize this function. The initial point is fixed to $(x,y)=(5,5)$. For Adam, we use the default setting that $\beta_1 = 0.9, \beta_2=0.99$. For FGD-MA and FGD-AR, we select $a_1 = b_1 = 0.9$ for order 1 model and $a_1 = b_1 = 0.8, a_2= b_2=0.1$ for order 2. For FGD-K, we select $\gamma=2,c=1,\sigma_Q^2=0.01,\sigma_R=2$. For FGD-W, we use Daubechies 4~(db4) wavelet and $K=3$ and $L=8$. We apply soft-thresholding on each highpass band with threshold of $0.2$ and select $\alpha=5$. Each Algorithm runs for $T = 500$ iterations.
\\ 
\textbf{Result}. Figure~\ref{fig:surface} visualizes the function value surface. We also show the trajectory of the optimization process for different methods in Figure~\ref{fig:traj0.01} and~\ref{fig:traj0.03}. Figure~\ref{0.01value} and~\ref{0.03value} are the plots of the function value along with the number of iterations. \textbf{First}, FGD resolves the local minimum problem by boost the lowpass band of the gradient. Under both learning rates, SGD was finally trapped in a local minimum. On the other hand, FGD-K and FGD-W is able to find the global solution constantly with smoothed gradient. This shows that, compared with vanilla SGD that only rely on local gradient, FGD-K and FGD-W can robustly solve the non-convex optimization problem with filtered gradients. \textbf{Second}, not all filters work the same good. In comparison to  FGD-K and FGD-W, the momentum method~(FGD-AR(1)) and FGD-AR(2) succeeded in finding the optimal solution when the learning rate is properly set~(0.01 in this experiment) but still fails when the learning rate increases to 0.03. FGD-MA model fails under both settings. The filter selection is important for a good convergence. \textbf{Surprisingly}, Adam, as the most popular method recently, performed extremely poor on both setting, mainly because the non-smoothing property of the optimization problem.

Figure~\ref{0.01x var}, \ref{0.01y var}, \ref{0.03x var} and \ref{0.03y var} illustrate the variance of the gradient for each parameter. SGD bounces around ravines, resulting in noisy gradients and large variance on both $x$ and $y$. The moving-average model introduces large variance to the gradient by blindly adding the past gradient to the current update. FGD-AR(1)~(momentum) and FGD-AR(2) relieve the variance problem when the learning rate is small but fails when a large learning rate is used. Adam succeeds in suppressing the gradient variance by keeping a running average of second-order momentum, but nearly stops learning from the 200-th iteration. It is clear that FGD-K and FGD-W are able to reduce the variance of the gradient compared with the momentum-based methods and vanilla SGD. This experiment shows promising results that FGD not only reduces the variance in gradient descent, but also help to get rid of the local minimum in the non-convex optimization problem. It encourages us to apply FGD to other optimization problems, like training neural networks.

\subsection{Study 2: Multi-layer perceptron}
We train a multi-layer perceptron (MLP) on MNIST~\cite{lecun2010mnist} dataset using (1) SGD (2) Adam~\cite{kingma2014adam} (3) FGD-MA with order 1 and 2 (4) FGD-AR with order 1~(vanilla momentum) and 2 (5) FGD-K and (6) FGD-W. We compare their (1) performance, measured by test accuracy (2) convergence rate, measured by training loss.\\ 
\textbf{Experiment setting}. We establish a 3-layer neural network with 784 input units, 10 hidden units, and 10 output units. All training images are augmented with zero-padding of 2 pixels and random cropped to $28\times 28$. Each image is flattened as a 1-dimension vector. We set the initial learning rate $\eta= \{0.1,0.01,0.001\}$ and mini-batch size $batch size= \{4,16,64\}$. The learning rate is multiplied by 0.7 every epoch. For Adam, we use the default setting that $\beta_1 = 0.9, \beta_2=0.99$. For FGD-MA and FGD-AR, we select $a_1 = b_1 = 0.9$ for order 1 model and $a_1 = b_1 = 0.8, a_2= b_2=0.1$ for order 2. For FGS-K, we select $\gamma=5,c=1,\sigma_Q^2=0.01,\sigma_R=2$. For FGD-W, we use Daubechies 4~(db4) wavelet and $K=3$ and $L=16$. We apply soft-thresholding on each highpass band with threshold of $0.2$ and select $\alpha=5$. We train the network for 10 epoch. The network is trained with a cross-entropy loss.
\\ 
\textbf{Result}. Figure~\ref{fig:MNIST training loss} compares the training loss under different learning rate and batch size setting on MNIST classification task. \textbf{First}, convergence is slower when the learning rate is small and batch size is large.
In such circumstances, filtering techniques accelerate the training by enhancing the low-frequency part of the gradient. For example, when $lr=0.001$ and $batchsize=64$, variants of FGD show faster convergence compared with vanilla SGD.
\textbf{Second}, a large learning rate and small batch size result in sharply jittering curves, which demonstrates large randomness in training. For example, Adam and FGD-AR(1)~(momentum) fails under the case with $lr=0.1$ and $batchsize=4$.
It again shows that the momentum method is very sensitive to the hyper-parameter setting. In such a case, FGD-K and FGD-MA successfully produce smoother curves and reduce the noise and randomness in the training. In comparison, FGD is more robust to learning rate changes compared to the momentum-based method.

Table~\ref{tab:MNIST} shows the test accuracy given various parameter setup on MNIST. \textbf{First}, FGD outperforms vanilla SGD when learning with small learning rate and large batch size. For example, FGD-W, FGD-MA, FGD-AR(1)\&(2) and FGD-MA(1)\&(2) ourperform SGD when $lr=0.01$ and $lr=0.001$ for all batch setting. FGD-W and FGD-AR even improves SGD by over 130\% for case where $lr=0.001$ and $batchsize=64$.
It reveals that FGD is able to accelerate learning and achieve better convergence when the step-size is small. \textbf{Another observation} from the table is that the momentum-based model~(FGD-AR(1) and Adam) is very sensitive to the hyper-parameter setting. Test accuracy with Adam dramatically drops to around 20\% for situations with a large learning rate~(0.1). In comparison, FGD-MA and FGD-K are robust to all kinds of hyper-parameter setting. This result further provides pieces of evidence that blind acceleration with the momentum method is not universally optimal. On the other hand, gradient estimation with filters is able to adaptively find a robust estimation of the true gradient regardless of parameter setup. \textbf{In addition}, FGD-MA with order 1 and order 2 makes the best performance on average. Compared with the traditional momentum method, it provides a cheap solution to training the neural network with great potential.

 \begin{figure*}[t]
    \centering
     \includegraphics[width = 0.93\linewidth]{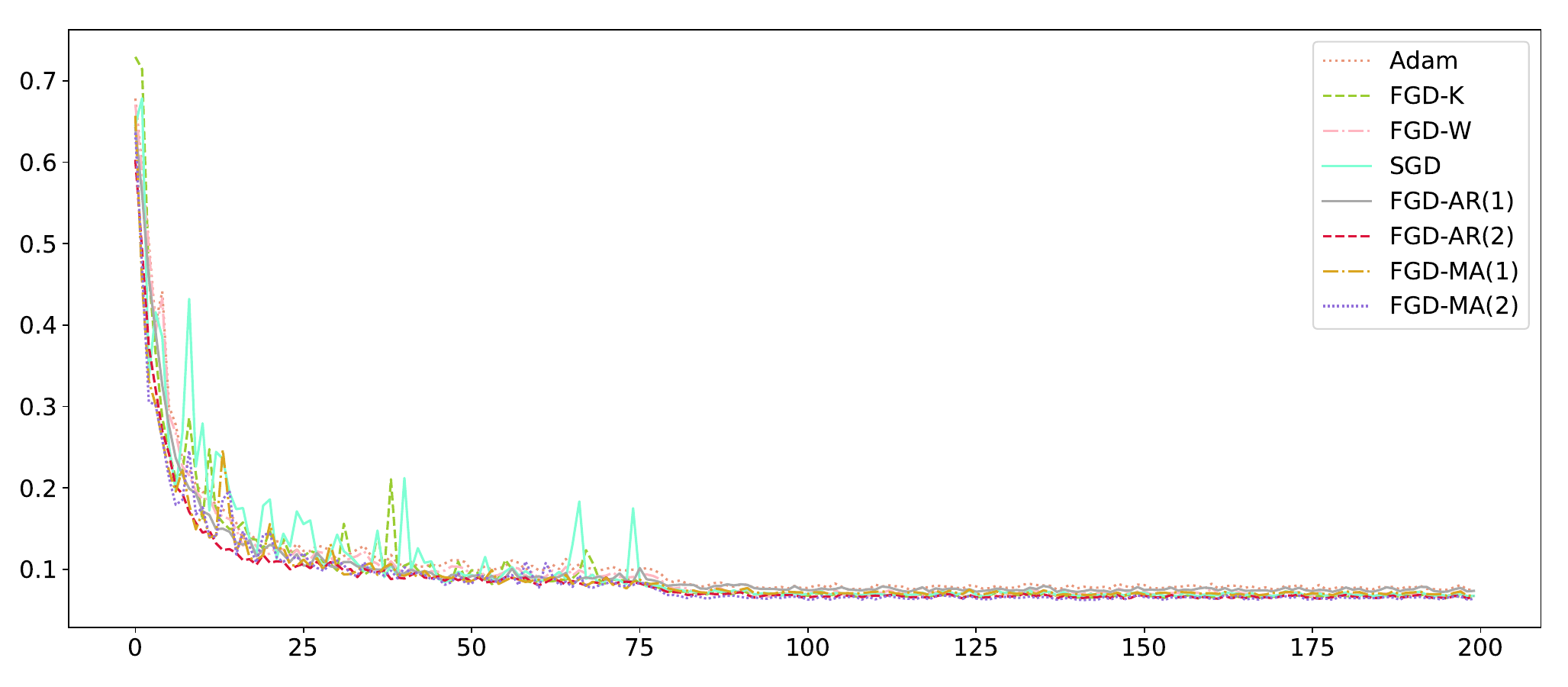}
     \includegraphics[width = 0.45 \linewidth]{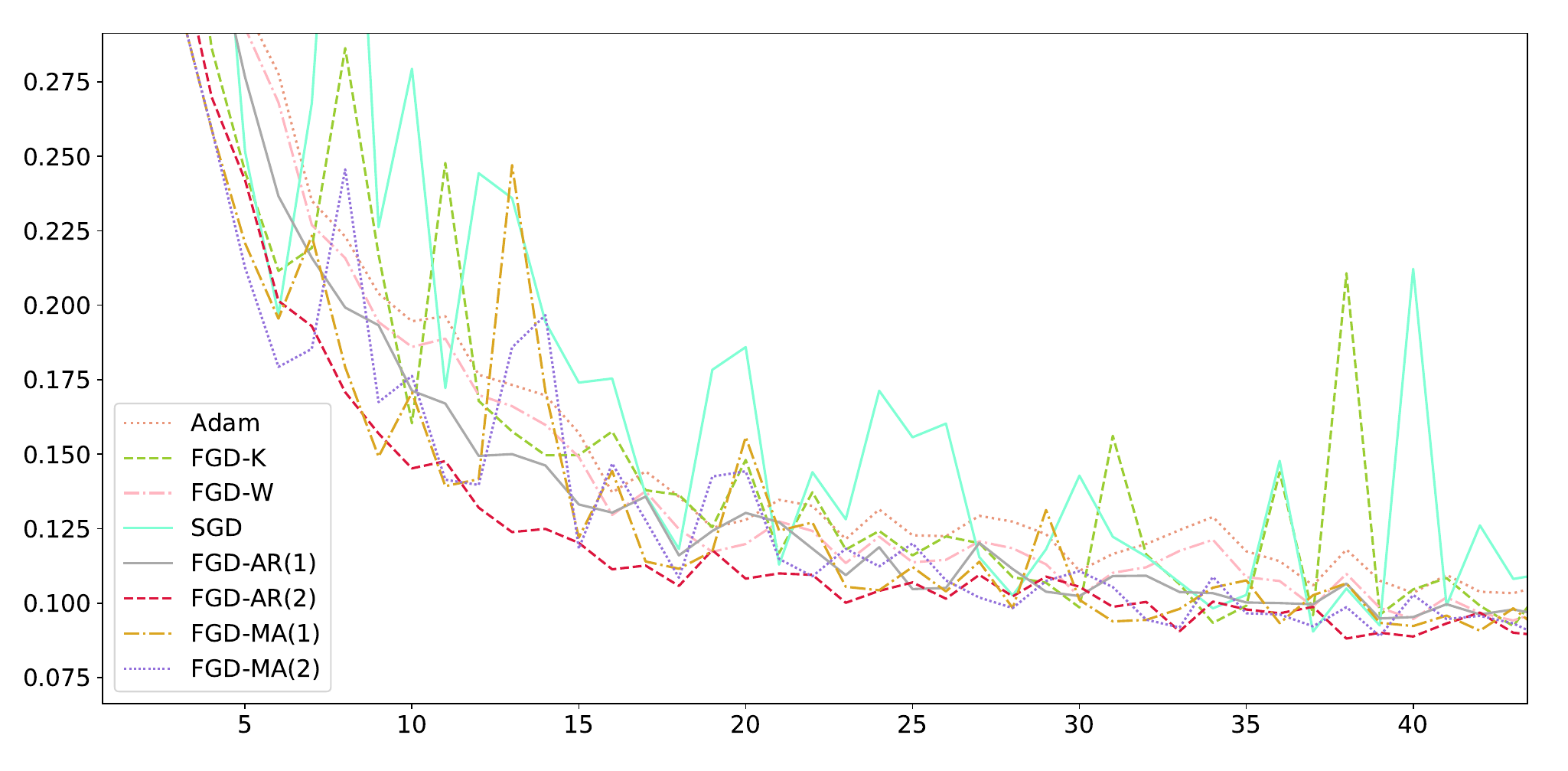}
     \includegraphics[width = 0.45 \linewidth]{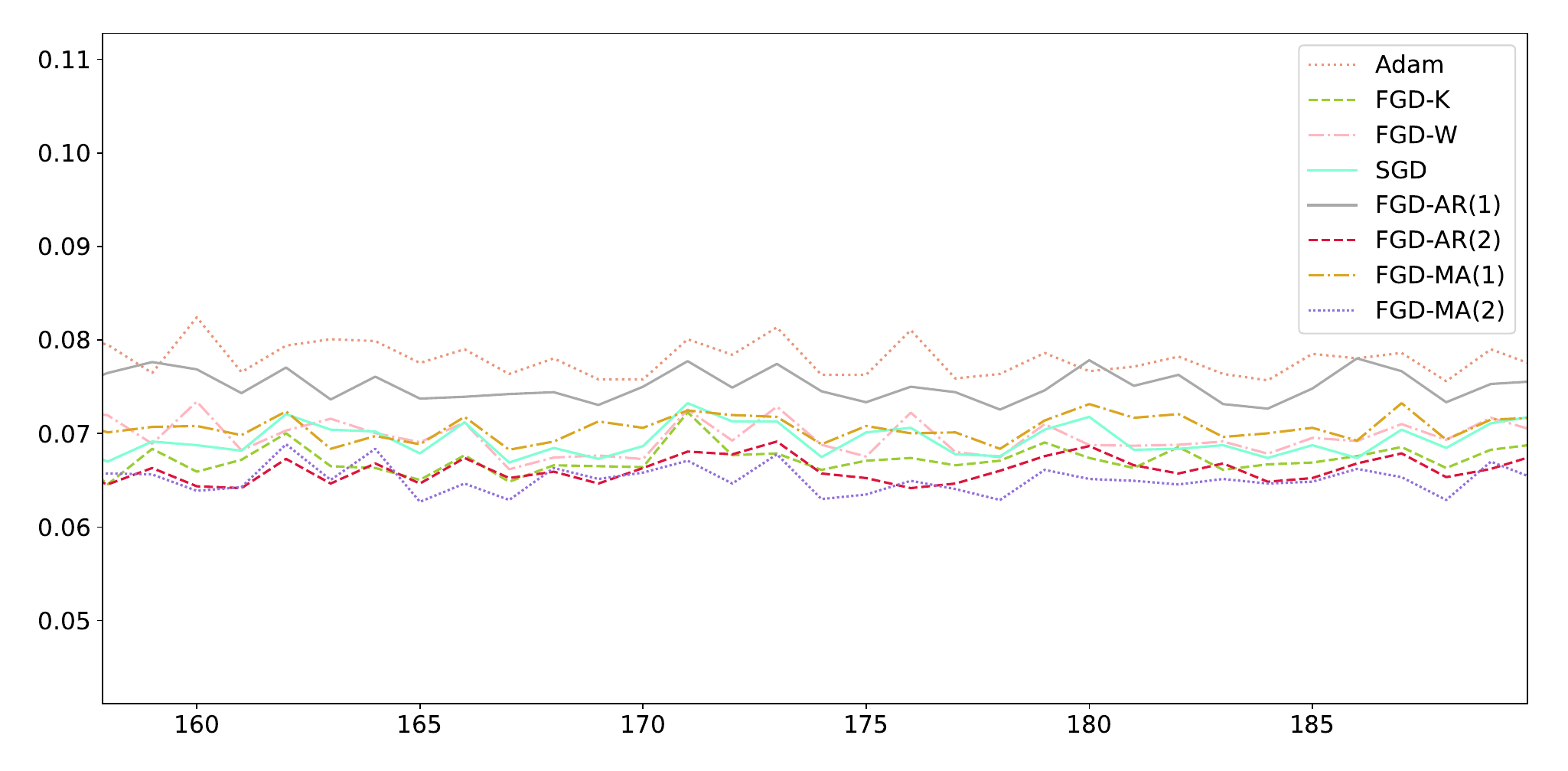}
    \caption{Test error curve On CIFAR-10 with ResNet-18, with two detail zoom in (x-axis: epoch; y-axis: test error)}
    \label{fig:Cifar acc}
\end{figure*}
\begin{table}[]
    \centering
    \caption{Test Accuracy on CIFAR-10 with ResNet-18}
    \begin{tabular}{l|c}
    \hline
         Method & Top1 test \\
         \hline
         Adam~\cite{kingma2014adam} & 92.50 \\
         FGD-AR(1) & 92.77\\
         SGD & 93.33\\\hline
         FGD-K & 93.54 \\
         FGD-M & 93.42\\
         FGD-AR(2) & 93.60\\
         FGD-MA(1) &93.23\\
         FGD-MA(2) &\textbf{93.76}\\
         \hline
    \end{tabular}
    \label{tab:cifar acc}
\end{table}
\subsection{Study 3: Convolutional neural networks}
We train a well-designed convolutional neural network (ResNet-18~\cite{resnet}) on CIFAR10~\cite{CIFAR-10} dataset using (1) SGD (2) Adam~\cite{kingma2014adam} (3) FGD-MA with order 1 and 2 (4) FGD-AR with order 1~(vanilla momentum) and 2 (5) FGD-K and (6) FGD-W. We compare their (1) convergence rate, measured by the test error curve, and (2) performance, measured by best test accuracy. 
\\ 
\textbf{Experiment setting}. All training images are augmented with random horizontal flip, zero-padding of 4 pixel and random cropped to $32\times 32$. We set the initial learning rate of $\eta= 0.1$, mini-batch size $batchsize= 64$ and train the network for 200 epoch. The learning rate is multiplied by 0.1 at the 80 and 160 epoch.  For Adam, we use the default setting that $\beta_1 = 0.9, \beta_2=0.99$. For FGD-MA and FGD-AR, we select $a_1 = b_1 = 0.9$ for order 1 model and $a_1 = b_1 = 0.8, a_2= b_2=0.1$ for order 2. For FGS-K, we select $\gamma=5,c=1,\sigma_Q^2=0.01,\sigma_R=2$. For FGD-W, we use Daubechies 4~(db4) wavelet and $K=3$ and $L=16$. We apply soft-thresholding on each highpass band with threshold of $0.2$ and select $\alpha=5$. The network is trained with a cross-entropy loss.
\\ 
\textbf{Result}. Table~\ref{tab:cifar acc} list the test accuracy for different optimization methods on the CIFAR-10 dataset. \textbf{First}, resembling with results in Study 2, momentum-based method~(Adam \& FGD-AR(1)) may not be optimal even compared with vanilla SGD. It again shows that blindly accelerating the gradient descent by using an $AR(1)$ model does not capture the real dependency of the noisy gradient observations and the true state. \textbf{Second}, FGD variants outperform SGD and momentum method experiment. These results reemphasize the major merit of the 
FGD method -- it is able to help the network to get rid of the global minimum while less sensitive to hyper-parameter selection. \textbf{Third}, we can make the observation that FGD is not sensitive to the filter selection. FGD achieves similar test accuracy regardless of which filter is applied. This result gives supporting evidence that FGD is a generally efficient framework in optimization. \textbf{Impressively}, among all variants, FGD-MA(2) achieves the best test performance of 93.76\% for ResNet-18 on the CIFAR-10 dataset. This result reemphasizes the huge potential for the FGD-MA method in training the neural network.

Figure~\ref{fig:Cifar acc} demonstrates the test error \textsc{VS} epoch through the training process. The bottom-left and bottom-right plots provide two zoom-in views for the training process. The bottom-left plot is in the middle of training and the bottom-right plot visualizes the test error at the end of the training. \textbf{First}, in the bottom-left plot, we may observe that the variance in the error curve is successfully reduced by the different filters compared with vanilla SGD~(light blue solid). The curve SGD is highly non-smooth, with tremendous ups and downs. Filtering in gradients, especially for FGD-AR(2) and FGD-K, succeed in reducing the variance of the gradient, resulting in smoother curves. \textbf{Second}, in the bottom-right plot, we see that FGD results in better convergence in training neural networks. Compared with Adam~(orange dotted) and FGD-AR(1)~(gray solid), FGDs generally achieve lower test error. For example, FGD-MA(2)~(purple dotted) and FGD-AR(2)~(crimson dashed) achieve the lowest test error. This observation illustrates that FGD is efficient in training the neural network.  

\section{Conclusion}

In this paper, we propose Filter Gradient Decent, which reduces the variance in gradient estimation by solving a filtering problem. It is a general framework that applies filter techniques to stochastic gradient descent to estimate the true state of the gradient from noisy observations. FGD not only greatly reduces the noise in the gradient estimation, but also helps to get rid of the local optima problem in the optimization. A series of comprehensive experiments show that FGD is able to achieve superior performance in solving non-convex optimization problems and training neural networks.

\section{Team Member Contribution}
Xingyi Yang: 100\%
\section*{Future work}
Filtering in signal processing is a well-studied subjects. A large number of researchers have provided mathematical proofs that different filtering techniques can efficiently eliminate different noise models. In future works, I would like to provide some theoretical provements that Filter Gradient Decent is mathematically explainable in reduce the gradient variance. 
\section*{Acknowledgment}
The authors would like to thank Professor Truong Nguyen for his quality curriculum design and careful explanation for the course content .Thanks for all the classmates in ECE251C. Hope everyone stay healthy.



%

\bibliographystyle{IEEEbib}
\bibliography{ref}

\begin{thebibliography}{10}

\bibitem{wang2013variance}
Chong Wang, Xi~Chen, Alexander~J Smola, and Eric~P Xing,
\newblock ``Variance reduction for stochastic gradient optimization,''
\newblock in {\em Advances in Neural Information Processing Systems}, 2013, pp.
  181--189.

\bibitem{qian1999momentum}
Ning Qian,
\newblock ``On the momentum term in gradient descent learning algorithms,''
\newblock {\em Neural networks}, vol. 12, no. 1, pp. 145--151, 1999.

\bibitem{duchi2011adaptive}
John Duchi, Elad Hazan, and Yoram Singer,
\newblock ``Adaptive subgradient methods for online learning and stochastic
  optimization,''
\newblock {\em Journal of machine learning research}, vol. 12, no. Jul, pp.
  2121--2159, 2011.

\bibitem{zeiler2012adadelta}
Matthew~D Zeiler,
\newblock ``Adadelta: an adaptive learning rate method,''
\newblock {\em arXiv preprint arXiv:1212.5701}, 2012.

\bibitem{hinton2012neural}
Geoffrey Hinton, Nitish Srivastava, and Kevin Swersky,
\newblock ``Neural networks for machine learning lecture 6a overview of
  mini-batch gradient descent,''
\newblock .

\bibitem{kingma2014adam}
Diederik~P Kingma and Jimmy Ba,
\newblock ``Adam: A method for stochastic optimization,''
\newblock {\em arXiv preprint arXiv:1412.6980}, 2014.

\bibitem{dozat2016incorporating}
Timothy Dozat,
\newblock ``Incorporating nesterov momentum into adam,''
\newblock 2016.

\bibitem{bottou2012stochastic}
L{\'e}on Bottou,
\newblock ``Stochastic gradient descent tricks,''
\newblock in {\em Neural networks: Tricks of the trade}, pp. 421--436.
  Springer, 2012.

\bibitem{schmidt2017minimizing}
Mark Schmidt, Nicolas Le~Roux, and Francis Bach,
\newblock ``Minimizing finite sums with the stochastic average gradient,''
\newblock {\em Mathematical Programming}, vol. 162, no. 1-2, pp. 83--112, 2017.

\bibitem{miller2017reducing}
Andrew Miller, Nick Foti, Alexander D'Amour, and Ryan~P Adams,
\newblock ``Reducing reparameterization gradient variance,''
\newblock in {\em Advances in Neural Information Processing Systems}, 2017, pp.
  3708--3718.

\bibitem{9414588}
Xingyi Yang,
\newblock ``Kalman optimizer for consistent gradient descent,''
\newblock in {\em ICASSP 2021 - 2021 IEEE International Conference on
  Acoustics, Speech and Signal Processing (ICASSP)}, 2021, pp. 3900--3904.

\bibitem{kalman1960new}
Rudolph~Emil Kalman,
\newblock ``A new approach to linear filtering and prediction problems,''
\newblock 1960.

\bibitem{lecun2010mnist}
Yann LeCun, Corinna Cortes, and CJ~Burges,
\newblock ``Mnist handwritten digit database,''
\newblock {\em ATT Labs [Online]. Available: http://yann. lecun.
  com/exdb/mnist}, vol. 2, 2010.

\bibitem{resnet}
Kaiming He, Xiangyu Zhang, Shaoqing Ren, and Jian Sun,
\newblock ``Deep residual learning for image recognition,''
\newblock {\em arXiv preprint arXiv:1512.03385}, 2015.

\bibitem{CIFAR-10}
Alex Krizhevsky, Vinod Nair, and Geoffrey Hinton,
\newblock ``Cifar-10 (canadian institute for advanced research),''
\newblock .

\end{thebibliography}

\end{document}